Original Paper

# A Fast and Minimal System to Identify Depression Using Smartphones: Explainable Machine Learning–Based Approach

Md Sabbir Ahmed, BSc; Nova Ahmed, PhD

Design Inclusion and Access Lab, North South University, Dhaka, Bangladesh

**Corresponding Author:**
Md Sabbir Ahmed, BSc
Design Inclusion and Access Lab
North South University
Plot #15, Block #B, Bashundhara
Dhaka, 1229
Bangladesh
Phone: 880 1781920068
Email: msg2sabbir@gmail.com

## Abstract

**Background:** Existing robust, pervasive device-based systems developed in recent years to detect depression require data collected over a long period and may not be effective in cases where early detection is crucial. Additionally, due to the requirement of running systems in the background for prolonged periods, existing systems can be resource inefficient. As a result, these systems can be infeasible in low-resource settings.

**Objective:** Our main objective was to develop a minimalistic system to identify depression using data retrieved in the fastest possible time. Another objective was to explain the machine learning (ML) models that were best for identifying depression.

**Methods:** We developed a fast tool that retrieves the past 7 days' app usage data in 1 second (mean 0.31, SD 1.10 seconds). A total of 100 students from Bangladesh participated in our study, and our tool collected their app usage data and responses to the Patient Health Questionnaire-9. To identify depressed and nondepressed students, we developed a diverse set of ML models: linear, tree-based, and neural network–based models. We selected important features using the stable approach, along with 3 main types of feature selection (FS) approaches: filter, wrapper, and embedded methods. We developed and validated the models using the nested cross-validation method. Additionally, we explained the best ML models through the Shapley additive explanations (SHAP) method.

**Results:** Leveraging only the app usage data retrieved in 1 second, our light gradient boosting machine model used the important features selected by the stable FS approach and correctly identified 82.4% (n=42) of depressed students (precision=75%, F1-score=78.5%). Moreover, after comprehensive exploration, we presented a parsimonious stacking model where around 5 features selected by the all-relevant FS approach Boruta were used in each iteration of validation and showed a maximum precision of 77.4% (balanced accuracy=77.9%). Feature importance analysis suggested app usage behavioral markers containing diurnal usage patterns as being more important than aggregated data-based markers. In addition, a SHAP analysis of our best models presented behavioral markers that were related to depression. For instance, students who were not depressed spent more time on education apps on weekdays, whereas those who were depressed used a higher number of photo and video apps and also had a higher deviation in using photo and video apps over the morning, afternoon, evening, and night time periods of the weekend.

**Conclusions:** Due to our system's fast and minimalistic nature, it may make a worthwhile contribution to identifying depression in underdeveloped and developing regions. In addition, our detailed discussion about the implication of our findings can facilitate the development of less resource-intensive systems to better understand students who are depressed and take steps for intervention.

*(JMIR Form Res 2023;7:e28848)*   doi: 10.2196/28848

## Introduction

**Background**

Depression is found in around 280 million people worldwide [1]. Although it is a common mental disorder, 80% of its burden is found in people from low- and middle-income countries (LMICs) [2]. It is the most prevalent mental disorder among adults in Bangladesh [3]. Moreover, the depression rate among Bangladeshi university students is higher than in other groups [4]. It is linked with physical illness [5] and also with psychological problems, such as anxiety disorder [6]. However, 75% of people living in LMICs do not receive any treatment for mental disorders [7]. In this case, social stigma is a barrier [7], which highlights the need for an unobtrusive way to identify depression. In contrast, there are people with psychological problems who seek support from primary care providers (PCPs). However, in more than 50% of cases, PCPs fail to recognize depression [8,9]. Failing to identify depressed individuals at an early stage may have devastating consequences as this increases the risk of suicide [10]. Recent research has shown that 60% of people who committed suicide struggled with major depression [11]. Therefore, there is a need to identify depression faster, which may make a significant contribution to mitigating depression through early intervention [12].

With the advent of computational models, extensive research has been conducted on the development of machine learning (ML) models for depression identification. There are subjective data–based studies in which to develop models, researchers used demographic characteristics [13-16], information about the family [14,15], lifestyle [15], mental health [13-17], etc, and their findings demonstrated good accuracy of the models. For instance, an ML model correctly identified 64% of depressed participants [14]. However, the main limitation of subjective studies is the use of self-reported data-based features for models, which makes the process obtrusive. Additionally, due to using features such as gender [14,15], which remains constant, the models may not be able to capture changes in depression over time, which in turn may not work for remote monitoring and faster diagnosis. Furthermore, since manual input is required, these approaches can have fewer implications in resource-constrained clinical settings.

To overcome these problems, researchers have explored pervasive devices leveraging behavioral data to improve mental health. Using Fitbit- and smartphone-sensed data, researchers [18] explored the behavioral patterns linked with loneliness and identified students with loneliness with an accuracy of 80.2%. A previous study [19] constructed 2 different data sets with behavioral data for over 100 days and predicted depression with an $F_1$-score of over 80%. Another study [20] presented a personal behavioral model to predict depression that correctly identified over 80% of depressed students using data from 10 and 16 weeks. Although wearable technology shows promising performance, these systems need to run for the whole period of data collection (eg, for over 100 days [19,21]). Thus, the need for a long data collection period may not facilitate early intervention. In addition, the high price of wearable devices can make them unaffordable for people of low income [22], which in turn may make the approaches infeasible for low-resource settings, where usage of wearable devices, such as Fitbit, is low.

Smartphones have become affordable [23] and are available to the majority of adults in emerging and developing countries [24]. Smartphone usage has a significant relationship with depression [25-28] and loneliness [29-31]. Moreover, there remain significantly different use patterns between depressed and nondepressed individuals in terms of communication [25] and social media [26] app categories, which indicates that app usage data can be important predictors for identifying depression. Based on only phone usage data, an ML model in previous research [21] showed a sensitivity of 45% in predicting postsemester depression, whereas another study [27] achieved a sensitivity of 55.7% in identifying participants with depressive symptoms. To develop ML models, some studies have used sensed data [19,21,27,32,33], along with smartphone usage data. In other studies [34-36], researchers have used smartphone-sensed data incorporating self-reported data to extract features for ML models. Studies have also relied solely on smartphone-sensed location data [37-40]. However, like studies based on wearables, in existing smartphone data–based studies, systems (ie, apps) need to run in the background for the whole data collection period, which may cause several problems. For example, due to running in the background for a long time (eg, 12 weeks [28], 16 weeks [19-21]) as well as sensors consuming too much battery power [41], these smartphone-sensed data-based systems may not be energy efficient, which can be a barrier to obtaining quality data from low-resource settings. Although some studies (eg, [40]) were conducted to develop energy-efficient systems for depression identification, the systems may still not facilitate early intervention since a long data collection period is required. In addition, due to having a system tracking data continuously from the phone, users may feel reluctant, which may introduce research reactivity problems (eg, the Hawthorne effect [42]), causing biases in app usage data.

**Objective**

To overcome the aforementioned limitations, such as the high price of wearables, the need to run the system in the background, and the need for long-term data, our primary objective was to develop a system, named *Mon Majhi* (in English, *Mind Navigator*), that can identify depression unobtrusively following a minimalistic approach and in real time. Another objective was to explain the best ML models that can facilitate a better understanding of depressed students among mental health care professionals and help them take steps in intervention.

## Methods

**A Tool to Retrieve App Usage Data Instantly**

*Development of a Data Collection Tool*

Self-reported app usage data do not represent actual behavior [43]. In reporting app usage, users overestimate or underestimate data, which varies by type [44]. For instance, Facebook usage duration is overestimated, while the frequency of launching Facebook is underestimated [44]. Therefore, to obtain the exact app usage behavioral data of users, we developed an app [45] for the Android platform, which is used by 95.9% of smartphone users in Bangladesh [46]. To instantly (see the *Time Required to Retrieve App Usage Data* section) obtain raw app usage data (foreground and background events), we used some functions of the Java class *UsageStatsManager*. However, app usage events are kept for a few days in Android [47], and thus, our tool can instantly retrieve app usage data for the previous 7 days, while aggregated app usage data (eg, total time spent on an app over 14 days) for longer than 7 days can be accessed by a Java application programming interface (API). Although the list of used apps can be accurately retrieved through the Java function *queryUsageStats (intervalType, beginTime, endTime)*, the usage duration data when the data collection period exceeded 7 days were not accurate, as we found by testing Mon Majhi in multiple ways. We tested the app by setting different values (eg, INTERVAL_WEEKLY, INTERVAL_MONTHLY) of the parameter *intervalType* as well as by properly changing the values of other parameters. By experimenting through the trial-and-error method, we found inaccuracy in the app usage duration when compared to the manually calculated app usage duration. In addition, while testing other phones, the users of the phones marked the data as inaccurate based on their best guess. Furthermore, except for usage duration, other data, such as frequency of launching, as well as raw data cannot be retrieved through the API, which hinders the extraction of more informative features, as presented in the *Pipeline of ML Models* section. Therefore, in this study, we used 7 days' app usage data as we found it to be accurate.

Once Mon Majhi (Figure A1 of Multimedia Appendix 1) is installed, with the user's consent, the app retrieves the past 7 days' app usage data. We chose to use Google Firebase as a database since it is secure and easy to integrate with mobile apps. We released the app [45] on the Google Play Store; since this platform is known to Android users, participants may feel more comfortable installing apps from there.

### Testing the Data Collection Tool

To check whether our app can accurately retrieve app usage data, we tested the app using the following 3 steps. We found that in each step, our app can calculate the past 7 days' app usage data, such as duration, and launch accurately.

- Step 1: We manually calculated the core app usage data duration and frequency of launch. We compared the manually calculated data with the app usage data retrieved by our data collection tool.
- Step 2: We compared the data retrieved by our app to the available apps [48,49] in Google Play that are required to be run in the background to calculate app usage data.
- Step 3: Smartphones function differently depending on various conditions, including the manufacturer and smartphone model. To determine the generalizable performance of our app, we tested it on 9 different smartphones.

### Time Required to Retrieve App Usage Data

To estimate the time required by our app to retrieve the past 7 days' app usage data, we calculated the time difference between the start and the end of programs that were written to retrieve app usage data. To estimate a generalizable required time, we tested our app on 20 smartphones of 19 different models, with 8 different versions of Android operating systems and 7 different smartphone manufacturers. Our app retrieved the past 7 days' app usage data 500 times from each of those phones. In total, it calculated the required time 10,000 (500×20) times. On average, it retrieved 7447.61 (SD 4986.62, median 6641, minimum=306, maximum=24,297) foreground and background app usage events (Figure 1a). The average time required was 307.94 ms (SD 1103.91, median 211, minimum=13, maximum=61,087 ms; Figure 1b). We found that among the 10,000 instances, only 97 had a retrieval time above 1 second (Table A1 of Multimedia Appendix 1).

In retrieving data, there may be some factors affecting the time required. We explored the possible factors by performing a correlation of the required time with the 20 phones' API level and the number of retrieval events. We did not find any significant correlation between the Android API level and the time required ($r_s$=0.18, $P$=.44; Figure 1c). In exploring the relationship with the number of events, we used the average number of events and the average time required for each phone; within each phone, there was almost no variation in the number of retrieved events (Figure 1a). We found the number of events had a significant positive relationship with the time required to retrieve data ($r_s$=0.56, $P$=.009; Figure 1d). Next, to estimate the plausible number of events in a student's phone, we used the data set that was constructed for this research. On average, there were 8174.04 events (SD 4972.50) retrieved from each of the 100 students' phones. In the 10,000 times data was retrieved, the number of retrieved app usage events was more than 8000 in the case of 4500 instances. To retrieve this large number of events, our app needed an average of 430.31 (SD 1596.46) ms. This reveals that, on average, our app can retrieve the past 7 days' app usage data in less than 1 second.

Figure 1. Performance of Mon Majhi in retrieving data. (a) Number of retrieved foreground and background events, (b) time required to retrieve data. Kernel density estimation shows the relationship of time with (c) API level and (d) number of foreground and background events. API: application programming interface.

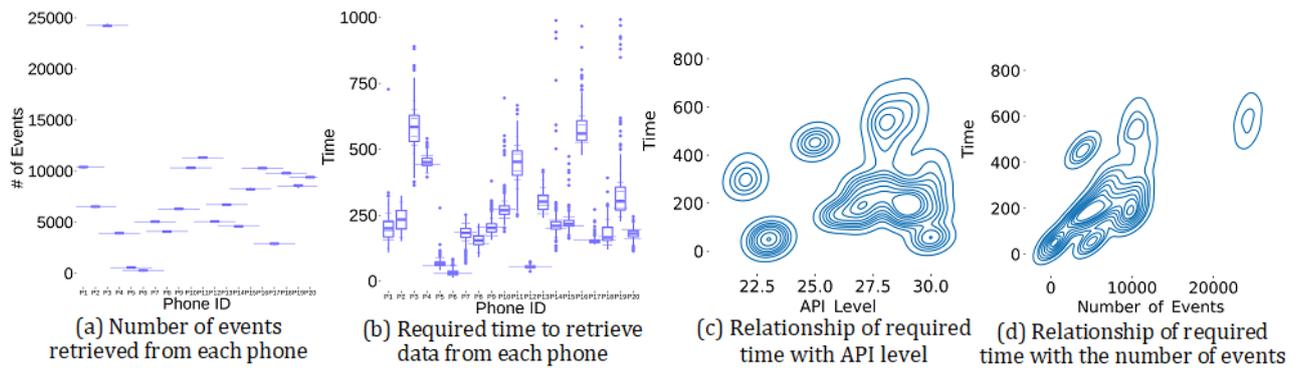

### Ethical Considerations

The study was approved by the Center for Research & Development at Eastern University. In our app, there was a consent form where we mentioned study details, and all participants provided their consent before their voluntary data donation. Except for the app usage data, our app does not collect any other data, such as messages. All data were kept anonymous, and access to the database was limited to the researchers of this project only.

The Center for Research & Development (CDR) at Eastern University is the only section regarding research at Eastern University that takes care of everything needed for research. There is no other separate section like an ethics board to review research.

### Data Collection and Participants' Demographic Characteristics

Considering the high prevalence of depression (ie, 69.5% [50]) among university students in Bangladesh as well as their high adoption of smartphones (86.6% use smartphones [51]), we decided to use them as samples in this study. We collected data from July to October 2020 and reached participants through the snowball sampling method. Several participants were recruited through university teachers, and others were recruited through researchers' close connections. To ensure the participants' comfort, we arranged a discussion session, where we described the study objective, types of data collected, etc. Since the study was conducted during the COVID-19 pandemic, we conducted the meeting mostly using virtual platforms based on participants' preferences and availability. To provide data, participants installed the app via Google Play. In total, 100 students from 12 different institutes of higher education and 7 different departments, including arts, law, medical science, and engineering faculties, participated.

The participants were from 36 districts and 7 divisions, which covered 56.3% of districts and 87.5% of divisions in Bangladesh (Figure 2a). There were 87 (87%) male participants and 13 (13%) female participants (Figure 2b). The participants' age varied from 19 to 30 years, and most participants' age was below 25 years (Figure 2c). Although a few participants had a family income of more than 100,000 Bangladeshi Taka (BDT) (US $942.32), most participants reported a family income of up to BDT 50,000 (US $471.16; Figure 2d).

Figure 2. Participants' demographic characteristics. (a) Pushpins present the location of districts, and bold text presents the division name. Participants' (b) gender, (c) age, and (d) monthly family income. BDT: Bangladeshi Taka.

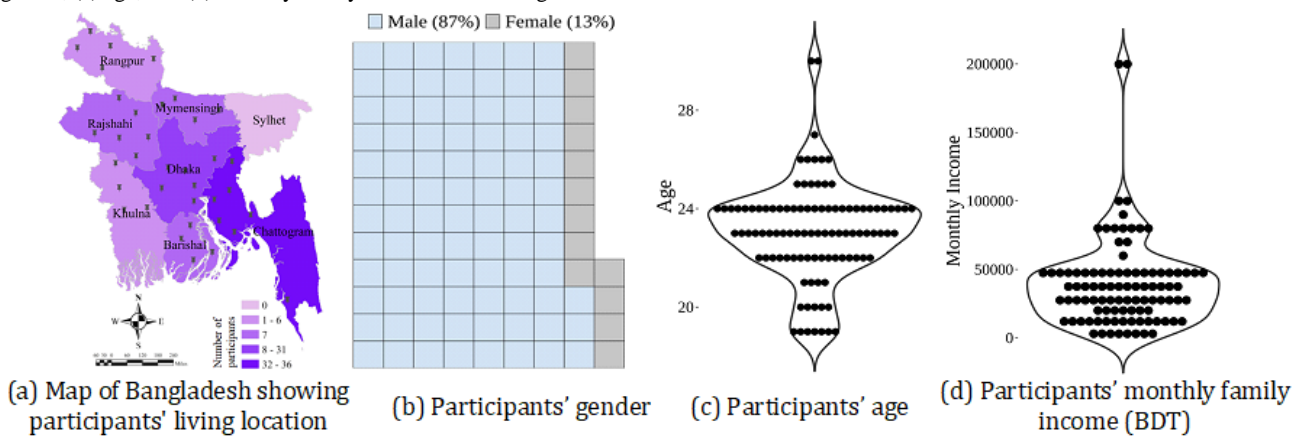

### Categorization of Depressed and Nondepressed Individuals

To assess depression among participants, different versions of the clinically validated Patient Health Questionnaire have been widely used [25,27,28,32,33,40,52]. We used the Patient Health Questionnaire-9 (PHQ-9) [53] in this study. In our app, the scale was available in English and in the native language Bengali (details about scale translation are available in section B of Multimedia Appendix 1). The PHQ-9 contains 9 items. Based on each participant's experiences in the past 14 days, they responded to each item using 1 of 4 options: not at all (0), several days (1), more than half of the days (2), and nearly every day (3). A PHQ-9 score of 10 or more showed a sensitivity and specificity of 88% for measuring major depression [53]. Therefore, following previous studies [25,26,28,53], we grouped participants who had a PHQ-9 score of at least 10 into the depressed group and those with a PHQ-9 score of <10 into the nondepressed group.

## Pipeline of ML Models

### Feature Extraction

From the 100 students' smartphones, our app retrieved 817,404 foreground and background events carried out in 1129 unique apps. The mean number of retrieved events from each phone was 8174.2 (SD 4972.53; median 7849; minimum=70, maximum=29,113). Using the retrieved data on foreground and background events, we extracted the frequency of launching apps, usage duration, and the following behavioral markers to use as features in the ML models: number of used unique apps, diurnal usage, app usage sessions, entropy, hamming distance, data of app categories, and number of extracted features.

### Number of Unique Apps Used

To count the number of unique apps used, we counted the number of app package names. The main motivation for using package names instead of app names was that among the 1129 apps (details are available in the *Usage Data of App Categories* section), 127 (11.2%) had duplicate names. For example, among the 8 apps in the Weather category, 7 (87.5%) app names were "Weather" and 1 (12.5%) app was named "Windy." However, each app's package name was unique.

### Diurnal Usage Data

The depressed and nondepressed groups had significantly different diurnal usage patterns [25,26]. Hence, we divided a day into 4 equal periods following previous studies [25,26]. During the calculation of app usage data, we found foreground and background events of several apps that occurred in different periods, and we added the duration in the respective period through the use of a delimiter. For example, if an app is opened at 11:30 AM (morning) but closed at 12:20 PM (afternoon), then by setting 12:00 PM (noon) as a delimiter, we added 30 minutes to the morning period and 20 minutes to the afternoon period. Moreover, following a previous study [54], in such cases, we counted the frequency of launch in the time range (eg, morning in this example) during which the app was opened:

- Night: 12:01 AM to 06:00 AM
- Morning: 06:01 AM to 12:00 PM
- Afternoon: 12:01 PM to 6:00 PM
- Evening: 06:01 PM to 12:00 AM

### App Usage Sessions

The Java function we used to retrieve the past 7 days' app usage data does not provide the phone lock and unlock data (eg, the time when the phone was locked or unlocked), which makes it difficult to identify a session. Hence, we followed previous studies to define a session. Wang and Mark [55] used the median break length of 40 seconds for grouping visits to Facebook into a single session. Other studies [56,57] have used a threshold of 60 seconds to identify a session of Facebook use. A threshold of 30 seconds has also been used to define the sequence of app usage into a single session [58-60]. However, using 30 seconds can identify sessions with less accuracy [61]. Instead, van Berkel et al [61] suggested using a threshold of 45 seconds, which was found to be more accurate. Therefore, in our study, we grouped app usage into a single session if there was no more than a 45-second gap between the last used app and the newly opened app. After that, depending on the time spent in each session, we defined 3 different types of sessions following previous studies [62,63]:

- Microsession: A session was defined as a microsession if a participant spent a maximum of 15 seconds on an app [63].
- Review session: A session was defined as a review session if a participant interacted with an app for up to 60 seconds [62]. However, due to the concept of microuse, we counted a session as a review session if a participant was found to spend between 15 seconds and 60 seconds on an app.
- Engage session: Following Banovic et al [62], we counted this session if the participant spent more than 60 seconds using apps on a smartphone.

### Entropy

Using Shannon's entropy formula [64], we calculated the entropy of every participant's app usage, which presents the app usage pattern:

$$E(i) = -\sum_{j=1}^{n} p(j) \log \log p(j),$$

where $p(j)$ indicates the probability of use of the $j$-th app by the $i$-th participant and

$$p(j) = \frac{usage_{duration}(j)}{\sum_{j=1}^{n} usage_{duration}(j)},$$

where $usage\_duration(j)$ presents the $i$-th participant's usage duration on the $j$-th app. Having an unequal usage duration on each app will result in lower entropy, $E$. In that case, the pattern of app usage will be skewed, and from that, we can infer that the participant has a preference for certain apps. If a participant uses a single app, the entropy ($E$) will be 0. Having an equal usage duration for every app will result in higher entropy.

### Hamming Distance Ratio

App signatures vary according to the group of people studied (eg, female vs male [65]). Depressed and nondepressed students have different app signatures, which makes them uniquely reidentifiable [26]. This difference is seen in app category as well [26]. Accordingly, we believe that uniqueness in terms of apps as measured by the hamming distance [66] can be a good metric to classify depressed and nondepressed students. For participant $i$, first, we calculated the distance from all the depressed participants:

$$D_{ij} = (AP_i \cup AP_j) - (AP_i \cap AP_j),$$

where $D_{ij}$ denotes the distance of the $i$-th participant from the $j$-th depressed participant and $AP_i$ and $AP_j$ denote the set of apps used by the $i$-th and $j$-th participants, respectively. Next, we found the minimum distance of participant $i$ from all ($n$) depressed participants: $D_i = min\{D_{i1}, D_{i2}, D_{i3}, \ldots, D_{in}\}$. Similarly, we calculated the minimum distance of participant $i$ in the nondepressed group, $ND_i$. After that, instead of using the distances ($D_i$, $ND_i$) separately as features, we used the ratio of the distances $RH_i = D_i / ND_i$. The motivation behind using the ratio is that it would provide us with information about how much more or less unique a participant is among the depressed group compared to the nondepressed group, and intuitively, this is more informative. Considering application in the real-world scenario where we have only app usage data, we did not use the information about the participant's (ie, participant $i$) category (depressed or nondepressed) during the calculation of the hamming distance ratio, which makes the feature unbiased.

*Usage Data of App Categories*

To calculate the usage data of an app category, we summed up the usage data of each app in that category. We took several steps while categorizing the apps. For instance, in Google Play, developers set the category of their app. For the apps used by the participants of our study, we retrieved the developers' referred category by using the app package name and an HTML parser. However, there were apps used by the participants that were not available in Google Play. To categorize those apps, we explored the app features from the online app stores (eg, APKMonk, APKMirror) and the developers' websites. For instance, participants used the Photo Editor app, available in the Samsung Galaxy Store, which we verified by matching the app's unique package name. After exploring the app, we found features (eg, adding effects on photos) regarding photography, and this directed us to keep the app in the Photo and Video app category. To categorize the apps, we also followed the app categorization process in previous studies [58,67]. In addition, we discussed this with 2 students who graduated from the computer science and engineering department. In the case of apps where there was disagreement among the categorizers, we discussed with 2 more students and used the majority rule to select a category. Due to having a small number of participants in each subcategory of the Games category, we grouped all the subcategories (eg, arcade, puzzle) into the Games category. In addition, since during the COVID-19 pandemic, students attended classes through apps, such as Zoom and Google Meet [68], we kept such apps in the Education category as all participants were students. After categorizing the 1129 apps the students used, we found that most (n=359, 31.8%) apps were in the Tools category and the least number of apps was in the Art and Design category (Table C1 of Multimedia Appendix 1). Moreover, we found more than 50 apps in the Games, Photo and Video, Books and Reference, Communication, and Productivity categories (Table C1 of Multimedia Appendix 1).

*Calculation of the Number of Extracted Features*

To simply show the calculation, we kept the data in different sets. A set of data presents the total smartphone usage (regardless of the app category) and 27 app categories: *app_category = {arts and design, . . . , weather, smartphone}*.

Since app usage behavior varies by weekdays and weekends [26,65], instead of using aggregated 7 days' (weekday + weekend) data, we used the weekday and weekend data separately as a feature. The set of the days was *days = {Weekdays, Weekends}*.

The sets of core data and session data were *core_data = {duration, launch, number of apps, entropy, hamming distance}* and *session_data = {total number of sessions, microsession, review session, engage session}*. We calculated the data for the whole day. In addition, we calculated the mean and SD for diurnal usage data consisting of morning, afternoon, evening, and night periods. We denoted these by the set of data characteristics: *data_characteristics = {mean, SD, total data}*.

In total, we extracted 864 features: (28 items in *app_category* × 2 items in *days* × 5 items in *core_data* × 3 items in *data_characteristics*) + (regardless of the app category: 2 items in *days* × 4 items in *session_data* × 3 items in *data_characteristics*). There were several app categories (eg, Art and Design, Auto and Vehicles) where the number of users was low. Having a nonuser creates a sparse matrix that may not demonstrate enough variance. We excluded all such features where the percentage of users was less than 50%. This resulted in 219 features (Table C2 of Multimedia Appendix 1) going through the feature selection (FS) step.

*Feature Selection*

Broadly, FS approaches are categorized into 3 groups: (1) wrapper, (2) filter, and (3) embedded methods [69]. We explored all 3 approaches. Moreover, we used the stable FS algorithm [70] as described later. In addition, to make the models unbiased to the features having larger values, we scaled the features where we performed standard scaling as some of the data contained outliers and standardization is less sensitive to outliers than min-max scaling [71].

*The Filter Method*

We used the information gain (IG) algorithm as the filter method. Unlike the Boruta algorithm, the IG algorithm does not inform a fixed set of features that can be optimal for classification. Hence, to select a set of top-scoring features, we set the lower bound by using the 1-in-10 rule [72] approach where the top 5 features were selected due to there being 51 depressed participants (see the *Depression Among Participants* section) in our study. Gradually, we increased the number of features to 20 to avoid the possibility of having an overfitted model with a large number of features.

*The Wrapper Method*

Unlike minimal-optimal methods, all-relevant features are selected in Boruta [73], where the random forest (RF) algorithm is wrapped. To implement this, we used the *BorutaPy* package [74], which works by correcting the *P* values in 2 steps rather than the 1-step Bonferroni correction, which is conservative. We changed the maximum depth of the RF from 3 to 7 [74], as suggested by the authors of the package.

### The Embedded Method

The embedded method combines the strategies of the filter and wrapper methods. We used the RF as the embedded method. In selecting the number of features based on the score of feature importance, we used the same approach as we did for the filter method.

### The Stable Method

In our study, we extracted 219 features from each of the 100 participants' app usage data. Due to a small number of participants, there may be unstable features where features may vary across different samples. The stable FS approach [70] was found to perform well in this scenario as many bootstrapped samples are created and the final set of features is selected based on a threshold ($\pi_{th}$), which presents the percentage of subsamples containing a feature. We created 1000 bootstrapped subsamples and used a logistic regression (Logit) classifier as the base estimator that fit on the bootstrapped subsamples. In previous studies of depression identification, researchers have used random thresholds (eg, 0.25 [75], 0.75 [76]) to select features. Since there is no evidence of getting optimal performance using only those thresholds, we performed an empirical investigation to present the optimal threshold. We started from a threshold of 0.5, which indicates that 50% of the bootstrapped subsamples contain a particular feature. Gradually, we increased the threshold by 0.01 up to a threshold where no more features were selected.

### Development and Validation of the Models

As there is no one-model-fits-all solution, we used a diverse set of ML algorithms, including those that are widely used in the medical informatics field, as shown in previous systematic reviews: decision tree (DT) [77], RF, support vector machine (SVM) [77-79], and Logit [78,79] algorithms. Moreover, we used other ML algorithms to increase the diversity of the models: Gaussian Naive Bayes, K-nearest neighbor (KNN), support vector classifier (SVC), AdaBoost, extra tree, multilayer perceptron (MLP) [71], light gradient boosting machine (LGBM) [80], CatBoost [81], and gradient boost (GB) [82]. As the baseline classifier, we used a dummy classifier. To develop and validate the ML models, we used the nested cross-validation (CV) method. In the outer loop was the leave-one-participant-out cross-validation (LOPOCV), and in the inner loop was a 20-fold CV, where 19 folds were used for tuning the hyperparameters and the remaining 1 fold was used for validation. LOPOCV maximizes the number of samples in training, where in each iteration, (N – 1) samples are used for training and 1 sample is used for testing (Figure 3). We took steps to prevent the possibility of overfitting the models. We used the nested CV method, which shows unbiased performance [83] and is used as the state-of-the-art method to restrain models from overfitting and overestimation [35]. Additionally, in the outer loop of nested CV, we used LOPOCV, which has a lower variance [84] and is used to minimize overfitting [21].

Hyperparameters in ML models play a role in enhancing performance. To tune the hyperparameters (the list of explored hyperparameters for the 13 ML algorithms is available in Table D1 of Multimedia Appendix 1), we used the Bayesian search optimization technique, which uses an informed search technique and works faster than the uninformed search technique (eg, grid search CV). It is worthwhile to mention that to develop unbiased ML models (Figure 3), the sample used in the testing was neither present in FS nor in the hyperparameter tuning steps. During model development, we maximized the $F_1$-score as it is based on the sensitivity and precision score, where sensitivity informed how many of the depressed participants were correctly classified and precision informed how many of the predicted depressed participants were truly depressed. After finding the best-performing models, we selected the top 5 algorithms as the base estimators to develop a stacking model, which works based on the wisdom of the crowd concept. To train the meta-learner Logit model of the stacking classifier, we used 10-fold CV. It is worthwhile to mention that in the case of each CV, we used the stratified technique so that the proportion of participants in each group remained the same in the training and testing parts, which ensured the unbiasedness of the model toward a particular group. For ML model development, we used Python packages, including *hyperopt* [85] and *sklearn* [86].

To evaluate the performance of the classification models, we used the evaluation metrics precision, $F_1$-score, and accuracy. However, an overfitted model can predict only a single class without being able to predict the other class. At that time, we will obtain ~50% accuracy in our data set as there was almost an equal number of participants in each group. Therefore, in addition to other evaluation metrics, to understand the performance in classifying the students in each group, we also calculated the sensitivity and specificity. Specificity informed us how many of the nondepressed participants were accurately classified.

**Figure 3.** Pipeline of Mon Majhi in identifying the depressed and nondepressed participants. DT: decision tree; GB: gradient boost; KNN: K-nearest neighbor; LGBM: light gradient boosting machine; LOPOCV: leave-one-participant-out cross-validation; MLP: multilayer perceptron; PHQ-9: Patient Health Questionnaire-9; RF: random forest; SVM: support vector machine; XGB: XGBoost.

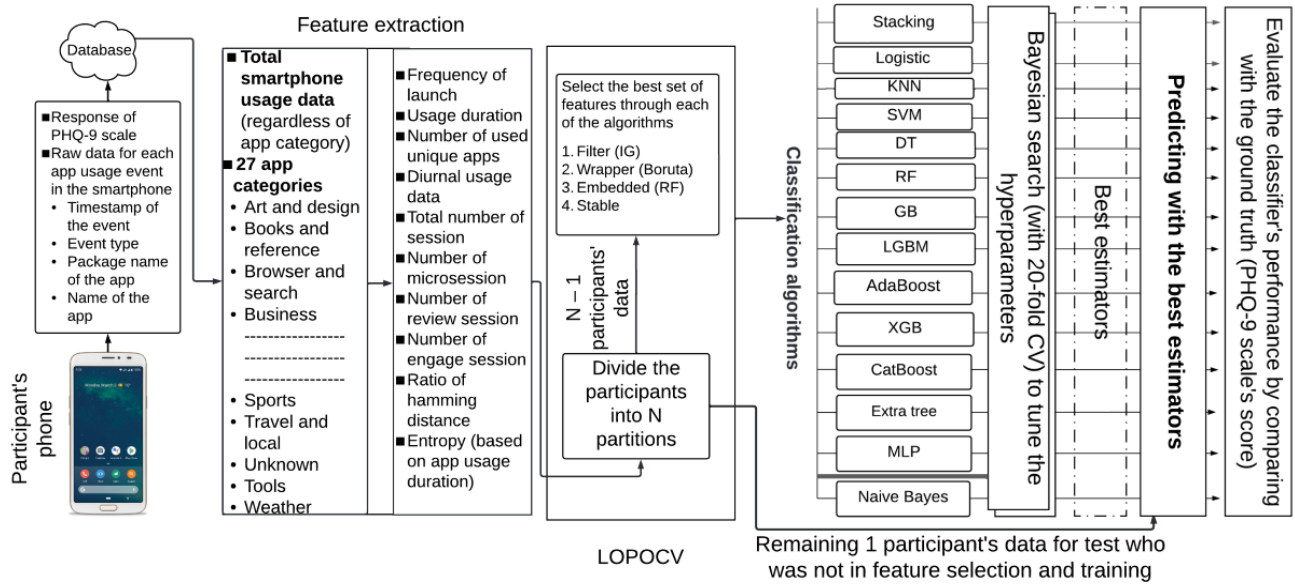

### Explanation of the Models

Explaining the models can be insightful for mental health professionals to understand depressed students. Additionally, this can help in the reproducibility of ML models. To understand how different features affect the probability of remaining in a particular group, we used the Shapley additive explanations (SHAP) [87] approach.

## Results

### Depression Among Participants

Of the 100 participants in our study, 51 (51%) had depression and 49 (49%) did not. The PHQ-9 score of the depressed group ranged from 10 to 27, whereas that of the nondepressed group varied from 1 to 9 (Figure 4a). After exploring differences in the 9 symptoms of the PHQ-9, we found that, on average, depressed students were bothered by each symptom for around more than half of the days in the past 14 days (score=2; Figure 4b). However, nondepressed students were not bothered by the symptoms at all, except symptom 1 (little interest or pleasure in doing things), where the average score was around 1 (several days; Figure 4b).

**Figure 4.** Depression score of the participants. (a) PHQ-9 score of the depressed and nondepressed participants. (b) Symptoms 1-9 of the PHQ-9 scale. Scores of 0, 1, 2, and 3 correspond to "not at all," "several days," "more than half of the days," and "nearly every day," respectively. PHQ-9: Patient Health Questionnaire-9.

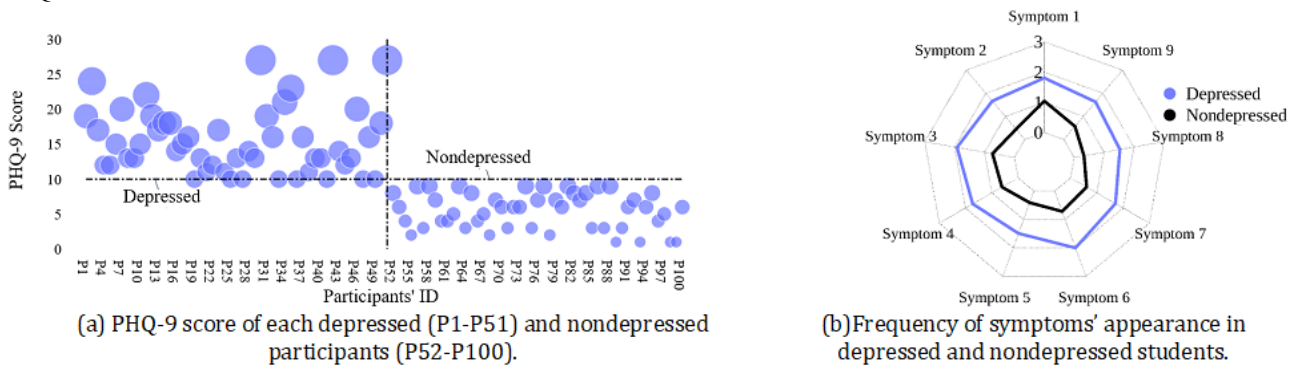

### Performance of the Models in Identifying Depression

In this section, we present the performance of ML models that are promising for identifying depressed and nondepressed participants. In Tables E1 to E4 of Multimedia Appendix 1, we present all 13 classifiers' performance in detail for models based on important features selected by the filter (IG), wrapper (Boruta), embedded (RF), and stable FS methods, respectively.

### Performance of ML Models in the Stable Feature Selection Approach

In finding the optimal threshold for the stable FS approach, we started from a threshold of 0.5 where a feature was selected if it was present in at least 50% of the 1000 bootstrapped subsamples. At a threshold of 0.5, on average, 61.9 features (SD 2.6) were selected in each iteration of LOPOCV (Figure 5a). We found the AdaBoost classifier performed the best in the selected features at this threshold, where precision, sensitivity, and specificity were 72.5%, 72.5%, and 71.4%, respectively (Figure 5b). When we increased the threshold by 0.01, gradually, the number of selected features decreased and reached 0 at a threshold of 0.98. At a threshold of 0.6, the number of selected features was 28.6, which was less than half of the number of features selected at a threshold of 0.5 (Figure 5a). However, from a threshold of 0.5 to 0.6, the models' performance did not vary largely (Figure 5b). In fact, the precision of the best models at each threshold was above 70%. In terms of precision, we found the best model at a threshold of 0.65, where the LGBM model–predicted depressed group was correct in 78% (n=39) of cases, and the sensitivity of 76.5% and specificity of 77.6% were also higher. Although an average of only 11.1 (SD 0.9) features were selected at a threshold of 0.77, in terms of sensitivity (82.4%) and the $F_1$-score (78.5%), the best performance was found at this threshold. However, the least number of features was selected at a threshold of 0.97, where in each iteration of LOPOCV, an average of 1.3 (SD 0.5) features were selected (Figure 5a) and the LGBM model's predictions were the most accurate (precision=63.3%, sensitivity=60.8%, specificity=63.3%; Figure 5b) at this threshold.

**Figure 5.** (a) Number of selected features and (b) performance of the best models at each threshold of the stable FS approach. The text at the end of the dotted lines presents the best models. FS: feature selection; Light GBM: light gradient boosting machine; LOPOCV: leave-one-participant-out cross-validation; ML: machine learning; MLP: multilayer perceptron.

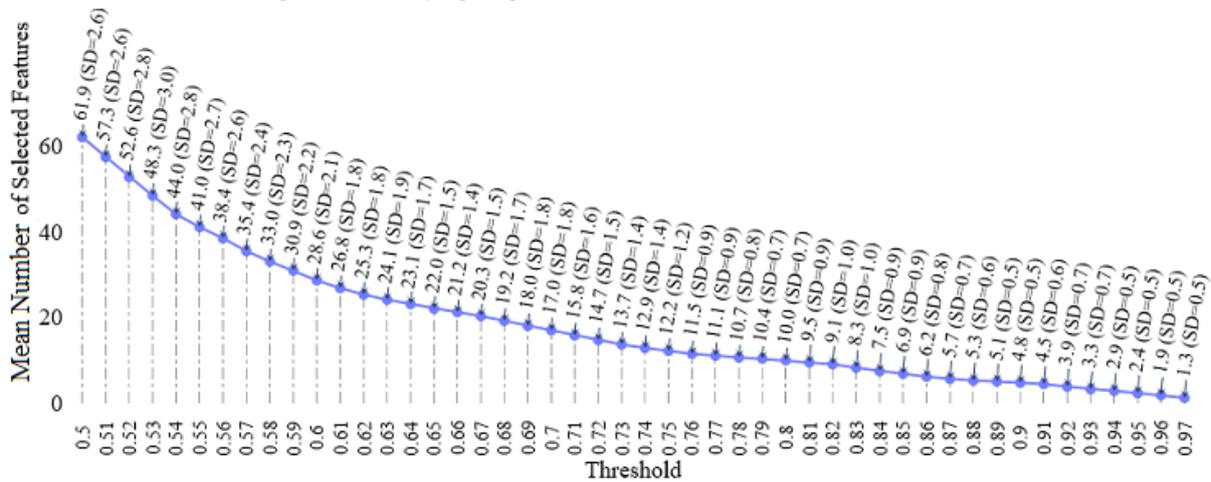

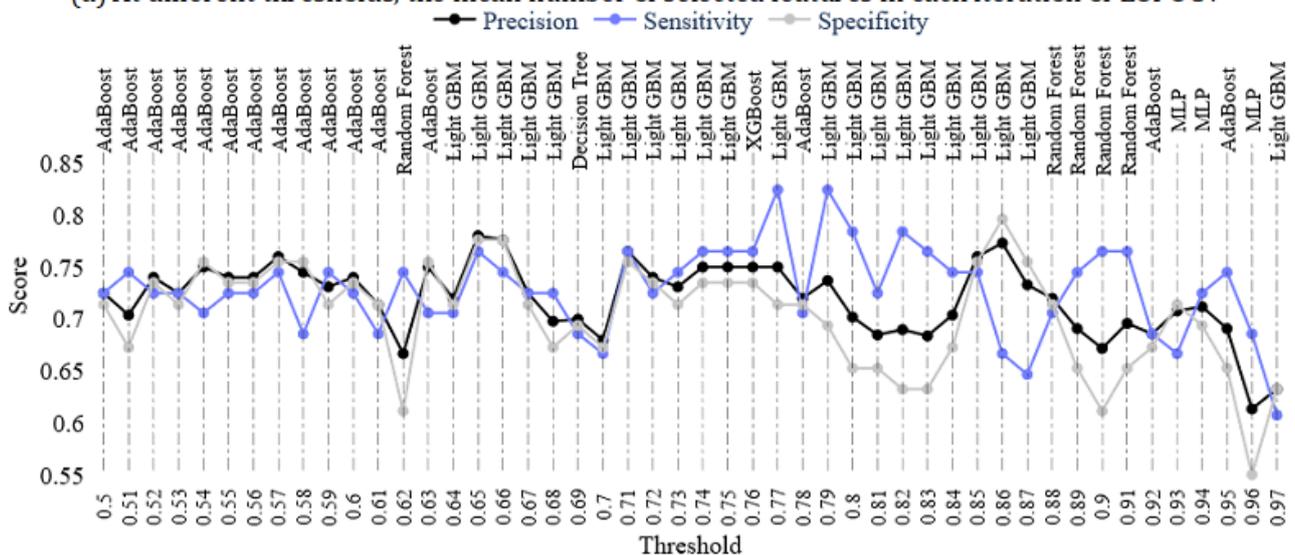

### Performance of ML Models in the Boruta Feature Selection Approach

Unlike all other FS approaches, in the wrapper method Boruta, we found all-relevant features showing higher performance for prediction tasks. To find the optimal performance of the models, we tuned the base estimator RF's maximum depth, which is wrapped in Boruta. On average, around 5 features were selected by Boruta when we varied the maximum depth from 3 to 7 (Figure 6a). Although the number of features did not vary, the set of selected features varied (Table F1 of Multimedia Appendix 1), which was reflected in the performance of the ML models (Figure 6b). We found the maximum sensitivity and $F_1$-score when depth was 4, where the KNN algorithm–based model showed a sensitivity of 82.4% and an $F_1$-score of 76.4%. However, the specificity (65.3%) of the model was below 70%. We found a better-performing model at a maximum depth of 6, where the GB algorithm model accurately identified 74.5% (n=38) of the depressed participants (sensitivity=74.5%), and the predicted depressed group was also correct in 73.1% (n=38) cases (precision=73.1%), with a specificity of 73.8% (Figure 6b).

**Figure 6.** (a) Number of selected features and (b) ML models' performance at different depths of the estimator of the Boruta algorithm. GB: gradient boost; KNN: K-nearest neighbor; LOPOCV: leave-one-participant-out cross-validation; ML: machine learning.

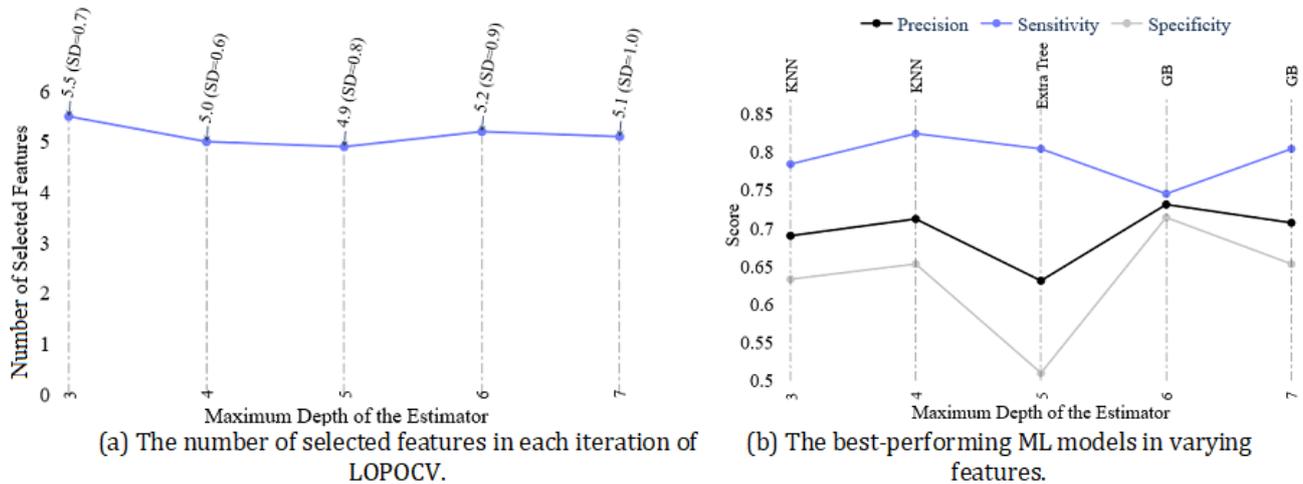

### Comparison of the Feature Selection Approaches

We compared the performance of the FS approaches by selecting the same number of features in each approach. In the filter method IG and the embedded method RF, we selected a subset of top-scoring features, where the lower and upper bounds were set following a process as described in the *Feature Selection* section. Unlike the filter and embedded methods, in the stable and also in the wrapper method Boruta, there was a variation in the number of selected features in each iteration of LOPOCV. Thus, for those 2 methods, we selected the best model by rounding the average number of selected features in LOPOCV so that the number of features becomes comparable. We found that using 5 features selected by the IG, the DT model performed better than the other 13 ML classifiers, where the precision, sensitivity, and specificity were 65.5%, 74.5%, and 69.7%, respectively (Table 1). Using the 5 features selected by the embedded and stable methods, CatBoost's (precision=64.9%, sensitivity=72.5%, specificity=59.2%) and the RF's (precision=69.1%, sensitivity=74.5%, specificity=65.3%) performance was higher, respectively (Table 2). However, the models based on the 5 features selected by Boruta outperformed the models based on the 5 features selected by the other 3 FS approaches (Tables 1 and 2). As mentioned previously, the GB model using 5 features selected by Boruta had a sensitivity, specificity, and $F_1$-score of 74.5%, 71.4%, and 73.8%, respectively (Table 1). In addition, it was interesting to observe that the same performance was found when using 6 and 7 features selected by the IG, 9 and 10 features selected by the RF, and 14 features selected by the stable FS approach (Tables 1 and 2). Thus, the GB model, developed by the 5 selected features of the all-relevant FS method, can be called the parsimonious model due to having a better predictive ability with a smaller number of features.

The optimal performance of the ML models varied by the number of selected features, as shown in Tables 1 and 2. For example, in the IG, the best performance was found using 9 features (DT: precision=76.6%, sensitivity=70.6%, specificity=77.6%), while in the RF FS approach, we found the best performance using 15 features (GB: precision=74.1%, sensitivity=78.4%, specificity=71.4%). Considering the performance of all models of all FS approaches, a model with the maximum $F_1$-score and sensitivity was found by using around 11 features of the stable FS approach (LGBM: precision=75%, sensitivity=82.4%, $F_1$-score=78.5%, specificity=71.4%). However, although the models' performance varied by the number of features, it appeared that in up to 10 features that were selected by each FS approach, there were several sets of features based on which the models' scores were around 70% when the sensitivity, specificity, precision, and $F_1$-score were calculated (Tables 1 and 2).

Table 1. Performance of the best models while selecting n features in the filter and wrapper FS[a] approaches.

| Features, n | Filter method (IG[b]) | | | | | Wrapper method (Boruta[c]) | | | | |
|---|---|---|---|---|---|---|---|---|---|---|
| | Best model | Precision | Sensitivity | $F_1$-score | Specificity | Best model | Precision | Sensitivity | $F_1$-score | Specificity |
| 5 | DT[d] | 0.655 | 0.745 | 0.697 | 0.592 | GB[e] | 0.731[f] | 0.745[f] | 0.738[f] | 0.714[f] |
| 6 | XGB[g] | 0.731[f] | 0.745[f] | 0.738[f] | 0.714[f] | KNN[h] | 0.69 | 0.784 | 0.734 | 0.633 |
| 7 | CatBoost | 0.731[f] | 0.745[f] | 0.738[f] | 0.714[f] | —[i,j] | — | — | — | — |
| 8 | DT | 0.750[k] | 0.706[k] | 0.727[k] | 0.755[k] | — | — | — | — | — |
| 9 | DT | 0.766[l] | 0.706[l] | 0.735[l] | 0.776[l] | — | — | — | — | — |
| 10 | LGBM[m] | 0.695 | 0.804 | 0.745 | 0.633 | — | — | — | — | — |
| 11 | LGBM | 0.712 | 0.725 | 0.718 | 0.694 | — | — | — | — | — |
| 12 | XGB | 0.691 | 0.745 | 0.717 | 0.653 | — | — | — | — | — |
| 13 | XGB | 0.702 | 0.784 | 0.741 | 0.653 | — | — | — | — | — |
| 14 | GB | 0.698 | 0.725 | 0.712 | 0.673 | — | — | — | — | — |
| 15 | LGBM | 0.714 | 0.686 | 0.700 | 0.714 | — | — | — | — | — |
| 16 | GB | 0.700 | 0.686 | 0.693 | 0.694 | — | — | — | — | — |
| 17 | DT | 0.778 | 0.686 | 0.729 | 0.796 | — | — | — | — | — |
| 18 | LGBM | 0.704 | 0.745 | 0.724 | 0.673 | — | — | — | — | — |
| 19 | XGB | 0.712 | 0.725 | 0.718 | 0.694 | — | — | — | — | — |
| 20 | XGB | 0.720 | 0.706 | 0.713 | 0.714 | — | — | — | — | — |

[a]FS: feature selection.

[b]IG: information gain.

[c]The number of features selected by Boruta and stable FS was rounded to make it comparable to the other FS approaches.

[d]DT: decision tree.

[e]GB: gradient boost.

[f]Low-performing (rank 3) classifiers in each approach.

[g]XGB: XGBoost.

[h]KNN: K-nearest neighbor.

[i]N/A: not applicable.

[j]In the Boruta method, the number of maximum important selected features was 5.5, so we set all values for 7-20 selected features as N/A.

[k]Medium-performing (rank 2) classifiers in each approach.

[l]High-performing (rank 1) classifiers in each approach.

[m]LGBM: light gradient boosting machine.

Table 2. Performance of the best models while selecting n features in the embedded and stable FS[a] approaches.

| Features, n | Embedded method (RF[b]) | | | | | Stable method[c] | | | | |
| --- | --- | --- | --- | --- | --- | --- | --- | --- | --- | --- |
| | Best model | Precision | Sensitivity | $F_1$-score | Specificity | Best model | Precision | Sensitivity | $F_1$-score | Specificity |
| 5 | CatBoost | 0.649 | 0.725 | 0.685 | 0.592 | RF | 0.691 | 0.745 | 0.717 | 0.653 |
| 6 | XGB[d] | 0.679 | 0.706 | 0.692 | 0.653 | LGBM[e] | 0.773 | 0.667 | 0.716 | 0.796 |
| 7 | GB[f] | 0.707 | 0.804 | 0.752 | 0.653 | LGBM | 0.760 | 0.745 | 0.752 | 0.755 |
| 8 | DT[g] | 0.735 | 0.706 | 0.720 | 0.735 | LGBM | 0.704 | 0.745 | 0.724 | 0.673 |
| 9 | LGBM | 0.731[h] | 0.745[h] | 0.738[h] | 0.714[h] | LGBM | 0.690 | 0.784 | 0.734 | 0.633 |
| 10 | LGBM | 0.731[h] | 0.745 | 0.738[h] | 0.714[h] | LGBM | 0.737 | 0.824 | 0.778 | 0.694 |
| 11 | LGBM | 0.755[i] | 0.725[i] | 0.740[i] | 0.755[i] | LGBM | 0.750[j] | 0.824[j] | 0.785[j] | 0.714[j] |
| 12 | LGBM | 0.755[i] | 0.725[i] | 0.740[i] | 0.755[i] | LGBM | 0.750[h] | 0.765[h] | 0.757[h] | 0.735[h] |
| 13 | DT | 0.735 | 0.706 | 0.720 | 0.735 | LGBM | 0.750[h] | 0.765[h] | 0.757[h] | 0.735[h] |
| 14 | KNN[k] | 0.720 | 0.706 | 0.713 | 0.714 | LGBM | 0.731 | 0.745 | 0.738 | 0.714 |
| 15 | GB | 0.741[j] | 0.784[j] | 0.762[j] | 0.714[j] | LGBM | 0.740 | 0.725 | 0.733 | 0.735 |
| 16 | LGBM | 0.712 | 0.725 | 0.718 | 0.694 | LGBM | 0.765[i] | 0.765[i] | 0.765[i] | 0.755[i] |
| 17 | RF | 0.696 | 0.765 | 0.729 | 0.653 | LGBM | 0.680 | 0.667 | 0.673 | 0.673 |
| 18 | DT | 0.660 | 0.686 | 0.673 | 0.633 | DT | 0.700 | 0.686 | 0.693 | 0.694 |
| 19 | LGBM | 0.729 | 0.686 | 0.707 | 0.735 | LGBM | 0.698 | 0.725 | 0.712 | 0.673 |
| 20 | GB | 0.712 | 0.725 | 0.718 | 0.694 | LGBM | 0.725 | 0.725 | 0.725 | 0.714 |

[a]FS: feature selection.

[b]RF: random forest.

[c]The number of features selected by Boruta and stable FS was rounded to make it comparable to the other FS approaches.

[d]XGB: XGBoost.

[e]LGBM: light gradient boosting machine.

[f]GB: gradient boost.

[g]DT: decision tree.

[h]Low-performing (rank 3) classifiers in each approach.

[i]Medium-performing (rank 2) classifiers in each approach.

[j]High-performing (rank 1) classifiers in each approach.

[k]KNN: K-nearest neighbor.

### Performance of the Stacking Models

After finding the optimal set of features for each FS approach, we built stacking models based on the top 5 classifiers. While selecting classifiers based on the embedded method's features, we found that although a model based on 15 features demonstrated the maximum $F_1$-score (Tables 1 and 2), most classifiers' performance remained higher while using 12 features (see Table 3; for comparison, see Table E3 of Multimedia Appendix 1). However, in the IG, Boruta, and stable methods, the best set of top 5 classifiers was found in 9 features, at the base estimator's maximum depth of 6, and at a threshold of 0.77, respectively (for details, see Tables E1, E2, and E4 of Multimedia Appendix 1).

Interestingly, from the top 5 classifiers of each FS approach, it was apparent that boosting models can find important behavioral patterns that make their predictions more accurate, keeping most of them in the top 5 classifiers list (Table 3), although we used a baseline dummy classifier and 13 different ML algorithms where linear and nonlinear algorithms were present. We found at least one variation of the GB models remained as one of the top 2 classifiers. The LGBM model in particular performed better across different sets of features. The LGBM model showed good performance consistently on most feature sets selected by the stable FS approach (Tables 1 and 2). Even when the number of features remained constant, the LGBM model remained one of the top 5 classifiers in each FS approach (Table 3).

The stacking models based on features selected by the filter, wrapper, and embedded methods had precision, sensitivity, specificity, and $F_1$-score values of more than 70%. The stacking model based on the features selected by the wrapper method Boruta correctly identified 80.4% (n=41; sensitivity=80.4%) of depressed participants. The predicted depressed group was also accurate in 77.4% (n=41) of cases (precision=77.4%), making this the most accurate model among all 4 stacking classifiers, as presented in Table 4. However, it was surprising to see that the stable method–selected feature-based stacking model had a precision and specificity lower than 70% (Table 4), although this method's selected feature set at a threshold of 0.77 produced the LGBM model with a higher sensitivity (82.4%) and $F_1$-score (78.5%) than any other individual model (Table 3). Additionally, we found the LGBM model had a relatively lower balanced accuracy— (sensitivity + specificity/2)—than the best stacking model: (82.4% + 71.4%)/2 = 76.9% for the LGBM model (Table 3) versus (80.4% + 75.5%)/2 = 77.9% for the Boruta-selected feature-based stacking model (Table 4).

**Table 3.** Top 5 classifiers and the baseline classifier's performance based on the performance of the best set of features of each FS[a] approach.

| Method and model name | Precision | Sensitivity | $F_1$-score | Specificity |
| --- | --- | --- | --- | --- |
| **Filter method (IG[b]; n=9 features)** | | | | |
| DT[c] | 0.766 | 0.706 | 0.735 | 0.776 |
| GB[d] | 0.717 | 0.745 | 0.731 | 0.694 |
| AdaBoost | 0.729 | 0.686 | 0.707 | 0.735 |
| LGBM[e] | 0.704 | 0.745 | 0.724 | 0.673 |
| XGB[f] | 0.692 | 0.706 | 0.699 | 0.673 |
| Baseline (dummy) | 0.510 | 1.000 | 0.675 | 0 |
| **Wrapper method (Boruta; base estimator's maximum depth=6)** | | | | |
| GB | 0.731 | 0.745 | 0.738 | 0.714 |
| KNN[g] | 0.707 | 0.804 | 0.752 | 0.653 |
| XGB | 0.725 | 0.725 | 0.725 | 0.714 |
| AdaBoost | 0.696 | 0.765 | 0.729 | 0.653 |
| LGBM | 0.714 | 0.686 | 0.700 | 0.714 |
| Baseline (dummy) | 0.510 | 1.000 | 0.675 | 0 |
| **Embedded method (RF[h]; n=12 features)** | | | | |
| GB | 0.732 | 0.804 | 0.766 | 0.694 |
| LGBM | 0.755 | 0.725 | 0.740 | 0.755 |
| Logit[i] | 0.750 | 0.706 | 0.727 | 0.755 |
| DT | 0.729 | 0.686 | 0.707 | 0.735 |
| KNN | 0.706 | 0.706 | 0.706 | 0.694 |
| Baseline (dummy) | 0.510 | 1.000 | 0.675 | 0 |
| **Stable method (threshold=0.77)** | | | | |
| LGBM | 0.750 | 0.824 | 0.785 | 0.714 |
| XGB | 0.745 | 0.745 | 0.745 | 0.735 |
| DT | 0.706 | 0.706 | 0.706 | 0.694 |
| GB | 0.706 | 0.706 | 0.706 | 0.694 |
| CatBoost | 0.691 | 0.745 | 0.717 | 0.653 |
| Baseline (dummy) | 0.510 | 1.000 | 0.675 | 0 |

[a]FS: feature selection.

[b]IG: information gain.

[c]DT: decision tree.

[d]GB: gradient boost.

[e]LGBM: light gradient boosting machine.

[f]XGB: XGBoost.

[g]KNN: K-nearest neighbor.

[h]RF: random forest.

[i]Logit: logistic regression.

Table 4. Performance of the stacking classifiers based on the top 5 classifiers of the best set of features of each FS[a] method.

| FS method | Precision | Sensitivity | $F_1$-score | Specificity | AUC[b] score | Accuracy |
| --- | --- | --- | --- | --- | --- | --- |
| Filter method (IG[c]; n=9 features) | 0.735 | 0.706 | 0.72 | 0.735 | 0.72 | 0.72 |
| Wrapper method (Boruta; maximum depth=6) | 0.774 | 0.804 | 0.788 | 0.755 | 0.78 | 0.78 |
| Embedded method (RF[d]; n=12 features) | 0.725 | 0.725 | 0.725 | 0.714 | 0.72 | 0.72 |
| Stable method (threshold=0.77) | 0.65 | 0.765 | 0.703 | 0.571 | 0.668 | 0.67 |

[a]FS: feature selection.

[b]AUC: area under the curve.

[c]IG: information gain.

[d]RF: random forest.

### Important Features and Explanation of the Models

Although each FS approach works differently, we found several common features as being important (Figure 7). There were 3 entropy-based features that were used more than 80% of the time among all iterations of LOPOCV in each of the 4 FS approaches. These features included two that measure the entropy based on the app usage of weekdays and weekends spanning a 24-hour period, as well as the feature that calculated the average entropy of weekdays during 4 time intervals (morning, afternoon, evening, and night; Figure 7). Although we used 12 different types of data, 40% (n=14) of the top 35 important features (all features are presented in Table F2 of Multimedia Appendix 1) were based on entropy, hamming distance, and session data (Figure 7). In fact, at a threshold of 0.97 of the stable FS approach, 98% of iterations of LOPOCV contained the *Weekday_Communication_Ratio_of_Hamming_6_Hour_Mean* feature (Table F3 of Multimedia Appendix 1), which is also based on hamming distance data. The hamming distance presents the app usage uniqueness, whereas entropy presents the app usage pattern, which decreases with higher inequality in app usage data. This reveals that complex app usage patterns can reflect better differentiable behavior, which can output higher classification accuracy.

In our extracted features, there were diurnal features presenting the app usage behavior in 6-hour intervals, as well as features based on the whole day. However, compared to the whole day's app usage, we found a larger number of important features regarding diurnal app usage behavior (n=19, 54.3%, for diurnal usage vs n=16, 45.7%, for 24-hour usage; Figure 7). In particular, a higher number of features (n=12, 34.3%) regarding the deviation of app usage behavior over the night, morning, afternoon, and evening periods was selected as important. We also found that although only 22.9% (n=8) of the top 35 features were based on overall smartphone usage data (regardless of the app category), 77.1% (n=27) were based on different app categories. We found the Communication, Social, and Tools app category–based features to be especially superior (Figure 7).

To explain the features' impact on the ML models' output, we used the SHAP approach. To check consistency, we explored the features' impact on the training as well as the test data for the best individual model (LGBM; Figure 8a,b) and the best stacking model (Figure 8c,d). In LOPOCV, (N – 1) participants' data were used for training and the remaining 1 participant's data were used for testing purposes. Therefore, a participant appeared *n* times during the training, whereas during the testing, a participant appeared only one time. This scenario is reflected in Figure 8, where there are more feature values in the summary plot based on training data. Interestingly, we observed consistency in the impact of the features on the model output. In the case of both training and testing data–based summary plots of the LGBM and stacking models, higher entropy in smartphone usage during the weekdays over a 24-hour period showed a negative impact (shifting the prediction toward the nondepressed group), while lower entropy showed a positive impact (moving toward the depressed group; Figure 8a-d). However, a higher mean entropy based on 4 time periods (ie, night, morning, afternoon, and evening) demonstrated an impact in the reverse direction (Figure 8a,b).

We also found in the Communication category that having a higher mean ratio of hamming distances in the 4 time periods during weekdays increased the predicted probability toward the depressed group, showing a positive impact (Figure 8a,b). Similarly, we found that higher Photo and Video app usage on the weekends moved the predicted probability toward the depressed group (Figure 8a,b). However, in the Education category, more time spent on weekdays appeared to increase the probability toward the nondepressed group (Figure 8a,b).

By local interpretations, we investigated how each participant's class probability was impacted by different features. As a sample, we presented 2 participants' group prediction approaches by the LGBM and stacking models, in which cases the prediction was accurate. We found that in the case of the nondepressed participant (ID 77), smartphone entropy based on a 24-hour period during the weekdays was 1.204 (Figure 9d), whereas in the case of the depressed participant (ID 46), this feature's value was –0.148 (Figure 9a), which was much lower. This finding indicates the same relationship as that presented in the summary plot in Figure 8. Moreover, the higher SD over the day in the number of photo and video apps in the case of depressed participants (Figure 9a) compared to nondepressed participants (Figure 9c) reflected the findings demonstrated in Figure 8, where we found the higher SD classifying the predicted group as depressed. From Figure 9, it is also apparent that to predict the group of this participant, a relatively smaller number of features were used in the Boruta-selected feature-based stacking model. For example, to predict the group of the depressed participant, the stable feature-based LGBM model used 8 features (Figure 9a), whereas for the same participant, the Boruta-selected feature-based model used only 5 features (Figure 9b), although both the models' predictions were correct.

**Figure 7.** Top 35 features among the features used for the best set of top 5 classifiers based on the filter method IG (n=9 features), wrapper method Boruta (base estimator's maximum depth=6), embedded method RF (n=12 features), and stable method (threshold=0.77). Here, features are ranked based on the mean appearance in the FS methods. The smartphone denotes data regardless of the app category. The values present the percentage of times (among all iterations of LOPOCV) a feature appeared. FS: feature selection; IG: information gain; LOPOCV: leave-one-participant-out cross-validation; RF: random forest.

| Feature | Filter | Wrapper | Embedded | Stable | Mean | Feature | Filter | Wrapper | Embedded | Stable | Mean |
|---|---|---|---|---|---|---|---|---|---|---|---|
| Weekday_Smartphone_Entropy_24_Hour | 99 | 100 | 100 | 100 | 99.8 | Weekday_Social_Duration_24_Hour | 0 | 1 | 55 | 0 | 14 |
| Weekend_Smartphone_Entropy_24_Hour | 100 | 95 | 100 | 100 | 98.8 | Weekday_Smartphone_Ratio_of_Hamming_24_Hour | 0 | 0 | 44 | 0 | 11 |
| Weekday_Smartphone_Entropy_6_Hour_Mean | 100 | 87 | 83 | 100 | 92.5 | Weekend_Browser_#_of_Apps_6_Hour_SD | 36 | 0 | 1 | 0 | 9.2 |
| Weekday_Communication_Ratio_of_Hamming_6_Hour_Mean | 0 | 98 | 96 | 100 | 73.5 | Weekend_Productivity_Launch_24_Hour | 29 | 0 | 0 | 0 | 7.2 |
| Weekday_Browser_Duration_6_Hour_SD | 100 | 34 | 84 | 0 | 54.5 | Weekend_Tools_Duration_24_Hour | 0 | 0 | 22 | 0 | 5.5 |
| Weekend_Smartphone_Ratio_of_Hamming_6_Hour_Mean | 0 | 9 | 99 | 97 | 51.2 | Weekday_Browser_Duration_24_Hour | 0 | 0 | 21 | 0 | 5.2 |
| Weekend_Productivity_Duration_24_Hour | 0 | 98 | 100 | 7 | 51.2 | Weekday_Communication_Launch_6_Hour_Mean | 0 | 0 | 19 | 0 | 4.8 |
| Weekday_Education_Duration_24_Hour | 0 | 0 | 66 | 100 | 41.5 | Weekday_Communication_Duration_6_Hour_Mean | 0 | 0 | 18 | 0 | 4.5 |
| Weekday_Tools_Entropy_24_Hour | 0 | 0 | 18 | 99 | 29.2 | Weekday_Photo_Video_Launch_6_Hour_SD | 0 | 0 | 16 | 1 | 4.2 |
| Weekend_Photo_Video_#_of_Apps_6_Hour_SD | 0 | 0 | 7 | 100 | 26.8 | Weekend_Tools_Duration_6_Hour_SD | 0 | 0 | 16 | 0 | 4 |
| Weekend_Communication_Entropy_6_Hour_SD | 100 | 0 | 4 | 0 | 26 | Weekday_Social_Launch_6_Hour_SD | 0 | 0 | 16 | 0 | 4 |
| Weekend_Photo_Video_#_of_Apps_24_Hour | 1 | 0 | 0 | 100 | 25.2 | Weekend_Tools_Ratio_of_Hamming_6_Hour_SD | 15 | 0 | 0 | 0 | 3.8 |
| Weekend_Smartphone_Review_Session_#_6_Hour_SD | 100 | 0 | 0 | 0 | 25 | Weekend_Productivity_#_of_Apps_24_Hour | 13 | 0 | 0 | 0 | 3.2 |
| Weekend_Social_Launch_6_Hour_SD | 0 | 0 | 0 | 98 | 24.5 | Weekend_Communication_Launch_24_Hour | 0 | 0 | 12 | 0 | 3 |
| Weekday_Smartphone_Entropy_6_Hour_SD | 67 | 1 | 15 | 0 | 20.8 | Weekday_Smartphone_Review_Session_#_6_Hour_Mean | 12 | 0 | 0 | 0 | 3 |
| Weekday_Social_Duration_6_Hour_Mean | 0 | 0 | 83 | 0 | 20.8 | Weekday_Communication_#_of_Apps_6_Hour_SD | 10 | 0 | 1 | 0 | 2.8 |
| Weekday_Photo_Video_Ratio_of_Hamming_24_Hour | 0 | 0 | 1 | 81 | 20.5 | Weekday_Social_Launch_24_Hour | 0 | 0 | 9 | 0 | 2.2 |
| Weekend_Productivity_Entropy_24_Hour | 60 | 0 | 0 | 0 | 15 | | | | | | |

**Figure 8.** Summary plot showing the impact of features on the output of the LGBM model and the stacking model: (a, c) training data and (b, d) testing data. Features are ranked by importance, which is calculated based on Shapley values. Here, we present the features that appeared in all iterations of LOPOCV. LGBM: light gradient boosting machine; LOPOCV: leave-one-participant-out cross-validation; SHAP: Shapley additive explanations.

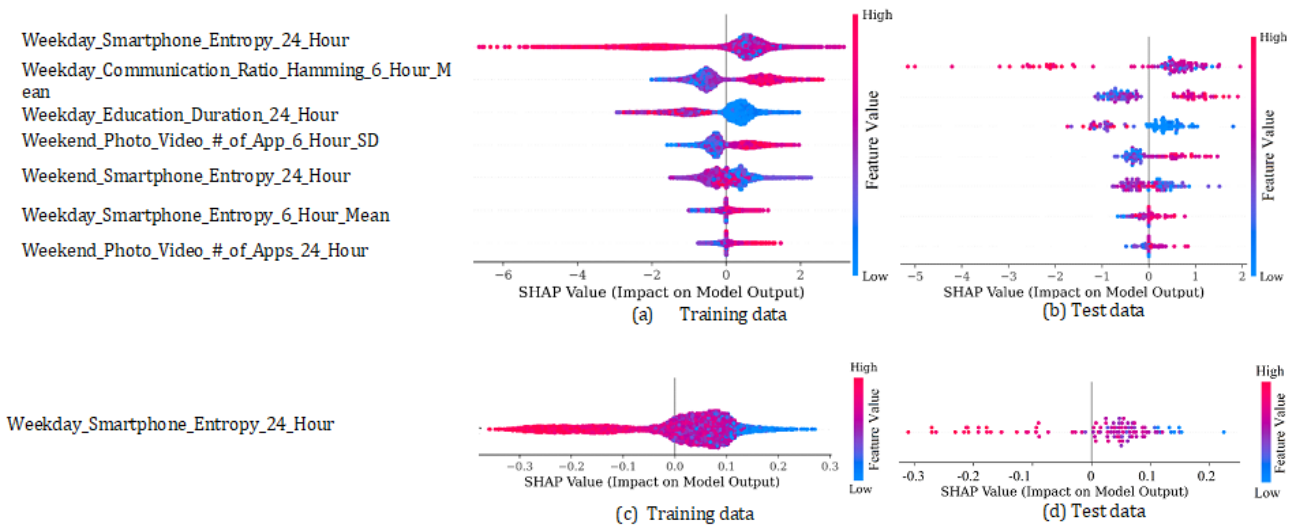

**Figure 9.** Force plot showing the identification of a participant with depression (participant ID 46) and a participant without depression (participant ID 77) by the LGBM (a, c) and stacking (b, d) models. Text in bold color shows the probability of remaining in the respected class (depressed, nondepressed). (a, b) Text in blue color shows the features moving the predicted class to 1 (depressed), while text in red color shows the features moving the predicted class to 0 (nondepressed). (c, d) The direction is reversed. Numerical values after each feature present the standardized feature value in the case of the participant that was used in model development. LGBM: light gradient boosting machine.

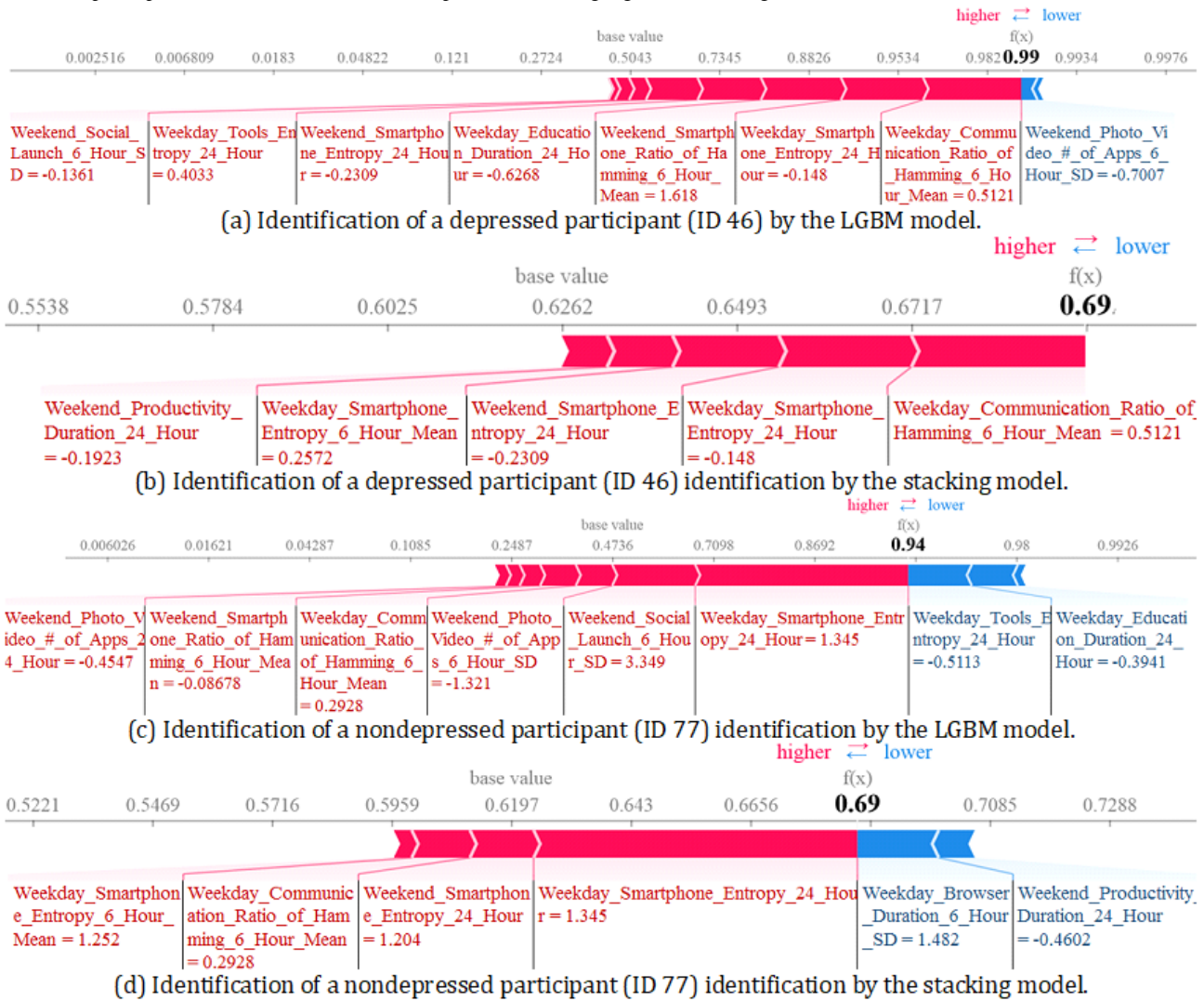

## Discussion

**Principal Findings and Comparison With Prior Work**

In this study, we presented Mon Majhi, a pervasive smartphone technology that aims to detect depression following a minimalistic approach in terms of data collection and detection time. It retrieves the past 7 days' app usage behavioral data within 1 second (mean 307.94, SD 1103.91 ms). Compared to the existing promising systems that leverage various data, including sensed data [19,20,28,32,36,87], phone usage [19,20,27,28,32,35,36], and network traffic [89], our system is faster and minimalistic, as presented in Table 5. For instance, an existing state-of-the-art systems [20] required 106 days to collect data of an equal number of days, and using those data, the model correctly identified 84.7% of depressed participants. However, using only our system's instantly (<1 second) accessed 7 days' data, our LGBM model correctly identified 82.4% of depressed participants. In addition, although previous studies have developed promising models to identify depressed participants, the systems in most studies rely on sensors, such as accelerometers [28,32], audio [28,32,88], Bluetooth [19,20,28], and GPS [19-21,28,32,36,88]. Using such sensors may be energy inefficient due to high power consumption (eg, by GPS [41]), which significantly reduces the battery life [41]. In addition, the need for current phone data–based systems [19-21,27,28,32,35,88,89] to run in the background may create reluctance as users want a long battery life [90] and running services in the background increases power consumption [91]. Moreover, students may not use systems for assessing mental health that negatively impact battery life [92]. Our system, on the other hand, does not use any such sensors or run in the background. We also do not need to run any resource-intensive systems (eg, a conversation classifier [32] or an audio signal processor [88]) to extract behavioral data. Instead, we perform simple mathematical calculations for feature extraction, which makes our system cheap, minimalistic, and largely scalable for resource-constrained settings. Our system can also be complementary to existing systems where faster detection can be followed by other systems for long-term evaluation. The minimalistic design opens opportunities for this system to be used in low-resource settings, such as an LMIC like Bangladesh.

Based on all evaluation metrics, our best stacking model had a performance of over 75%, while the robust models in a study [35] based on phone usage and internet connectivity had a performance of over 85% with a maximum accuracy of 98.1%. However, in addition to the general limitations discussed before, their models were limited by several other factors. For example, when classifying responses and considering that approximately 50% of participants had more than one response, it becomes possible for the same participants' responses to appear in both the training and testing data sets. This introduces the potential for biased models and overestimated performance [93]. Similarly, compared to a study by Leigh et al [27] where researchers developed ML models solely based on phone usage data, our stacking model demonstrated 18.4% higher sensitivity than their best model. There can be several reasons behind this enhanced performance. For instance, our feature importance analysis showed that a higher number of features pertaining to diurnal usage data are more important than the 24-hour aggregated data, which was unexplored by Leigh et al [27]. In the SHAP analysis, we found that a higher entropy based on smartphone usage in 24 hours of weekdays increased the model's predicted probability toward the nondepressed group, while an increase in average entropy in the 4 time periods of a day increased the predicted probability toward the depressed group. This also indicates that the diurnal usage pattern is different from the whole day's behavior. Additionally, in our study, features regarding entropy, hamming distance, and session data appeared as important, which also remained unexplored in the previous study [27]. At a threshold of 0.77 of the stable FS approach, a feature regarding the hamming distance appeared in each iteration of LOPOCV, where we found the best individual model: LGBM (precision=77%, sensitivity=82.4%). In addition, we found that a higher number of features based on app categories was important compared to the aggregated data regardless of app category. Behavioral markers of the particular app categories are associated with depression [25,26] and also with the rhythmic patterns of our body [95], as presented by researchers through conventional statistical methods [25,26,94] and also by a qualitative study [94]. Therefore, our findings suggest that while developing ML models, instead of leveraging only the aggregated phone usage data, as in the previous studies [19-21,27,28,32,35], incorporation of the features regarding app categories and extraction of features such as hamming distance may improve the performance of the ML models.

Table 5. Comparison of our system's performance with that of previous studies using pervasive devices to identify depressed and nondepressed participants. Researchers used binary classification in the included studies. The time required to collect data in each study based on the description of the data collection tool is reported.

| Reference and country (if available) | Sample size for models, N | Collected data | System needs to run in the background | Duration of explored data | Time to retrieve data | Accuracy | Precision | Sensitivity | Specificity | AUC[a] score | $F_1$-score |
|---|---|---|---|---|---|---|---|---|---|---|---|
| **Xu et al [19], United States** | | | | | | | | | | | |
| Data set 1 | 138 | Fitbit sensed, phone sensed, usage | Yes | 106 days | 106 days | 0.807 and 0.818 | 0.765 and 0.843 | 0.886 and 0.843 | —[b] | — | 0.821 and 0.843 |
| Data set 2 | 212 | Fitbit sensed, phone sensed, usage | Yes | 113 days | 113 days | 0.689-0.840 | 0.757-0.877 | 0.779-0.907 | — | — | 0.768-0.881 |
| **Xu et al [20], United States** | | | | | | | | | | | |
| Data set 1 | 138 | Fitbit sensed, phone sensed, usage | Yes | 106 days | 106 days | 0.825 | 0.862 | 0.847 | — | — | 0.855 |
| Data set 2 | 169 | Fitbit sensed, phone sensed, usage | Yes | 166 days | 166 days | 0.791 | 0.814 | 0.854 | — | — | 0.833 |
| Wang et al [32], United States | 83 | Microsoft Band sensed, phone sensed, usage | Yes | 63 days (9-week terms) | 63 days | — | 0.691 | 0.815 | — | 0.809 | — |
| Saeb et al [27], United States | 21 (for phone usage data-based analysis) | Phone usage | Yes | 14 days | 14 days | Mean 0.742 (SD 0.034) | — | Mean 0.640 | Mean 0.839 | — | — |
| Nickels et al [28], United States | 186 and 197 | Phone sensed, usage | Yes | 84 days | 84 days | — | — | — | — | Mean 0.620 (SD 0.062) and mean 0.656 (SD 0.079) | — |
| Opoku et al [35], mostly developed countries | 629 | Phone usage, internet, demography | Yes | Mean 22.1 (SD 17.9) days | Mean 22.1 (SD 17.9) days | 0.964-0.981 | 0.856-0.925 | 0.922-0.956 | — | 0.947-0.991 | 0.887-0.940 |
| Opoku et al [36] | 54 | Demographics, Oura ring sensed, phone sensed, usage | Yes | Mean 28.21 days | Mean 28.21 days | 0.814 | Dep[c]: 0.6997 Non-dep[d]: 0.841 | 0.505 | 0.924 | 0.823 | Dep: 0.587 Non-dep: 0.880 |
| **Dogrucu et al [88]** | | | | | | | | | | | |
| PHQ-9[e] cut-off score=10 and undersampling | 294 | Contacts, GPS, call log, social media, voice recording | Yes | 14 days | — | 0.588 | 0.599 | 0.554 | 0.623 | — | 0.575 |
| PHQ-9 cut-off score=20 and undersampling | 96 | Contacts, GPS, call log, social media, voice recording | Yes | 14 days | — | 0.771 | 0.783 | 0.75 | 0.792 | — | 0.766 |
| **Yue et al [89], United States** | | | | | | | | | | | |

| Reference and country (if available) | Sample size for models, N | Collected data | System needs to run in the background | Duration of explored data | Time to retrieve data | Accuracy | Precision | Sensitivity | Specificity | AUC[a] score | $F_1$-score |
|---|---|---|---|---|---|---|---|---|---|---|---|
| iOS users | 40 | Network traffic | Yes | Several months | Several months | — | 0.71 | 0.71 | 0.63 | — | 0.71 |
| Android users | 13 | Network traffic, screen on-off | Yes | Several months | Several months | — | 0.75 | 0.86 | 0.77 | — | 0.80 |
| **This study, Bangladesh** | | | | | | | | | | | |
| Performance of our best single classifier–based model: LGBM[f] | 100 | Phone usage data | No | Past 7 days | Mean 307.94 (SD 110.91) ms | .770 | 0.750 | 0.824 | 0.714 | 0.769 | 0.785 |
| Performance of our best stacking model based on the top 5 classifiers | 100 | Phone usage data | No | Past 7 days | Mean 307.94 (SD 110.91) ms | 0.780 | 0.774 | 0.804 | 0.755 | 0.780 | 0.788 |

[a]AUC: area under the curve.

[b]Not available.

[c]Dep: depressed.

[d]Nondep: nondepressed.

[e]PHQ-9: Patient Health Questionnaire-9.

[f]LGBM: light gradient boosting machine.

## Implications of Study Findings

In the stable FS approach, starting from a threshold of 0.5, we gradually increased the threshold by 0.01 until there remained 0 features and we found the best model in selecting the features that appeared at least 77% of the time among the 1000 bootstrapped subsamples. This finding highlights the need for empirical investigation of the threshold while using the stable FS approach. This finding also extends previous studies that have used random thresholds of 0.25 [75] and 0.75 [76] to select features for depression identification and also in other contexts (eg, Ing et al [95] used a threshold of 0.90 in neurobehavioral symptom identification). While comparing FS approaches, we found that to achieve the same performance (precision=73.1%, sensitivity=74.5%) as an ML model developed using around 5 features selected by Boruta, we need 6, 9, and ~14 features of the filter, wrapper, and stable approaches, respectively. In fact, using those 5 features of Boruta, our stacking model performed the best (precision=77.4%, balanced accuracy=77.95%). Unlike other FS approaches, selecting all-relevant features in Boruta [73], instead of selecting minimal-optimal features, can be a plausible reason for having better performance. Higher performance with a relatively lower number of features demonstrates the development of a parsimonious model that can have potential for resource-constrained settings where the usage of a higher number of features can increase the computational models' complexity and use relatively more resources. Additionally, the parsimonious model, based on all-relevant features in particular, can have potential implications in presenting plausible behavioral markers for intervention.

After comparing the performance of the ML models, we found the GB-based models GB, XGB, and LGBM to be superior, although we developed models using a baseline dummy classifier and 13 different classification algorithms, including the support vector, KNN, Logit, and neural network (MLP) algorithms, where each model was developed based on features selected by 4 different FS methods. In the GB ML models, the weak learners are converted to strong learners by correcting the predecessors through the gradient descent algorithm [82], and GB can effectively handle a complex relationship with nonlinearity [96]. This is reflected in the findings of our study, where we found better performance among GB-based models compared to linear models, such as Logit. In particular, we found the LGBM to be one of the top 5 classifiers among models based on features selected by each of the 4 methods. In fact, the best individual ML model was based on the LGBM algorithm, as mentioned before. A plausible reason for having better performance is the leafwise growth of the LGBM, which makes the model complex and also increases the complex relation-learning capability. Although complexity increases the possibility to have an overfitted model, we used the nested CV method, which is used for overfitting prevention and has an unbiased performance [83]. Our findings suggest that while using behavioral data for developing ML models to identify depression, GB-based ML algorithms, particularly the LGBM, may be a preferable choice, considering their lower consumption of memory and higher speed in computation [80].

While explaining the LGBM model through the SHAP method [87], it appeared that the greater the time spent on apps in the Education category during the weekday, the lower the probability of being depressed. In this study, apps including Zoom, Google Meet, and Google Classroom were in this category, and in Bangladesh, these apps have been used for online learning since the start of the COVID-19 pandemic [68]. Using these apps, students can support and communicate with their classmates and teachers. The positive impact of peer support on mental health [97] explains the plausible reason for the association with a lower probability of being depressed. A SHAP analysis of the model also showed that the higher number of photo and video apps used and also the higher deviation over the day in terms of the number of apps used in this category on the weekends increase the predictive probability toward the depressed group. The photos of depressed users are different, where their photos appear to be grayer, bluer, and darker [98]. In addition, studies have shown that those with mental health problems post photos on social media [99] and watch videos on YouTube [100] to share thoughts and seek support to overcome their problems. Therefore, as different apps have different features, the usage of a higher number of photo and video apps by depressed individuals can present their support-seeking behavior, which they prefer to do in a particular period of the day as there is a higher deviation over the day. Extending the previous studies [99,100], these findings show that in addition to visual attributes in social media, the behavioral features regarding the number of photo and video apps can also distinguish people who are or are not depressed. Meanwhile, in the case of the Communication category, we found that a higher ratio of hamming distances in communication apps is likely to increase the probability of being depressed, which denotes that depressed individuals are more likely to use a higher number of different communication apps. This finding is in line with a previous study [26] conducted using the conventional statistical method, which demonstrated the diverse nature of app usage among depressed people presenting their support-seeking behavior. Thus, going beyond depression identification, these explanations through SHAP analysis have the potential to help mental health care professionals better understand depressed individuals and take steps for intervention accordingly.

## Strengths and Limitations

Through our developed system Mon Majhi, we have contributed to the mobile and ubiquitous health research area in the following ways:

- Using only data retrieved in 1 second (mean 307.94, SD 1103.91 ms), our ML model correctly identified 82.4% of depressed individuals. To the best of our knowledge, in identifying a psychological problem, our approach is faster and more minimalistic than any other existing smartphone data–based systems, which can enable our system to be largely scalable in resource-constrained settings, such as in LMICs.

- We presented important behavioral markers and the best ML models after selecting features via 3 main types of FS approaches along with the stable approach and also after developing ML models based on 13 different classification algorithms. Due to the large exploration, our findings can have real-world implications. In addition, after a comprehensive exploration, we presented a parsimonious model based on features selected by the all-relevant FS method Boruta, which showed better predictability with a lower number of features. This can have potential for future studies to develop parsimonious computational models to identify psychological problems in low-resource settings leveraging behavioral data.

- Through explainable ML techniques, we interpreted the models where we demonstrated how different behavioral features impact predicting depression and also discussed the implications that can have potential for understanding the smartphone usage behavior of depressed students and in taking steps for intervention.

Our study was limited by the small sample size (N=100). Although there was diversity among the participants in terms of regions, institutions, and departments, and we used state-of-the-art methods to evaluate ML models, due to having a small sample size comprising mostly male participants, evaluation of a large sample is needed before applying the system in the real world. Additionally, while translating the PHQ-9 [53], we removed the word "dead" from the ninth item (eg, considering the students' concerns about the word "dead" [92]) through a process that is described in detail in section B of Multimedia Appendix 1. It should be noted that even after we removed the ninth item completely and considered the cutoff score of the 8-item Patient Health Questionnaire-8 (score ≥10: depressed) [101], all the depressed participants were still categorized as depressed. It is worthwhile to mention that since we constructed the data set of this study amid the COVID-19 pandemic in 2020 when classes were online, it was difficult to reach out to a large number of participants. However, we have been conducting a countrywide study where we have overcome the aforementioned limitations and constructed a large-scale data set. In our future work, we expect to present a more robust system to the research community.

## Conclusion

The performance of our system Mon Majhi showed that depressed and nondepressed students can be classified accurately, faster, and unobtrusively with minimal data. Although we developed models using a diverse set of ML algorithms, we found that the LGBM model using only instantly accessed data (<1 second) can correctly identify 82.4% of depressed students, with a precision of 75%. Additionally, we found the all-relevant FS approach Boruta-based stacking model (sensitivity=80.4%, precision=77.4%) as a parsimonious model due to higher performance with a lower number of features. Through a SHAP analysis, we also demonstrated how different app usage behavioral markers impact the models. These findings are novel and show the feasibility of our minimal system for faster depression prediction. Hence, we believe that our system can facilitate minimization of depression rates in low-resource settings.

## Acknowledgments

We thank all participants who voluntarily donated data. We are grateful to Tanvir Hasan and the faculties of the various universities that supported us in the data collection phase.

All authors declared that they had insufficient or no funding to support open access publication of this manuscript, including from affiliated organizations or institutions, funding agencies, or other organizations. JMIR Publications provided article processing fee (APF) support for the publication of this article.

## Conflicts of Interest

None declared.

## Multimedia Appendix 1

Description and analyses.
[DOCX File , 267 KB-Multimedia Appendix 1]

## Abbreviations

**API:** application programming interface
**BDT:** Bangladeshi Taka
**CV:** cross-validation
**DT:** decision tree
**FS:** feature selection
**GB:** gradient boost
**IG:** information gain
**KNN:** K-nearest neighbor
**LGBM:** light gradient boosting machine
**LMIC:** low- and middle-income country
**Logit:** logistic regression
**LOPOCV:** leave-one-participant-out cross-validation
**ML:** machine learning
**MLP:** multilayer perceptron
**PCP:** primary care provider
**PHQ-9:** Patient Health Questionnaire-9
**RF:** random forest
**SHAP:** Shapley additive explanations
**XGB:** XGBoost